\documentclass{article}

\usepackage{multirow}
\usepackage{arxiv}
\usepackage{float} 
\usepackage[utf8]{inputenc} 
\usepackage[T1]{fontenc}    
\usepackage{hyperref}       
\usepackage{url}            
\usepackage{booktabs}       
\usepackage{amsfonts}       
\usepackage{nicefrac}       
\usepackage{microtype}      
\usepackage{lipsum}		
\usepackage{graphicx}
\usepackage{natbib}
\usepackage{doi}
\usepackage{tabularx} 
\usepackage{booktabs} 
\usepackage{algorithm}
\usepackage{algpseudocode}
\usepackage{amsmath}

\title{Habaek: High-Performance Water Segmentation through Dataset Expansion and Inductive Bias Optimization}

\author{Hanseon Joo \and Eunji Lee\and Minjong Cheon\thanks{Corresponding author. Email: \texttt{jmj541826@gmail.com}}}

\date{
    \textsuperscript{1}Hanyang University, Seoul, South Korea.  \\
    \textsuperscript{2}Korea Racing Authority, Gwacheon, South Korea. \\
    \textsuperscript{3}Independent Researcher
}

\renewcommand{\shorttitle}{\textit{arXiv} Template}

\hypersetup{
pdftitle={A template for the arxiv style},
pdfsubject={q-bio.NC, q-bio.QM},
pdfauthor={Minjong},
pdfkeywords={First keyword, Second keyword, More},
}

\begin{document}
\maketitle

\begin{abstract}
Water segmentation is critical to disaster response and water resource management. Authorities may employ high-resolution photography to monitor rivers, lakes, and reservoirs, allowing for more proactive management in agriculture, industry, and conservation. Deep learning has improved flood monitoring by allowing models like CNNs, U-Nets, and transformers to handle large volumes of satellite and aerial data. However, these models usually have significant processing requirements, limiting their usage in real-time applications. This research proposes upgrading the SegFormer model for water segmentation by data augmentation with datasets such as ADE20K and RIWA to boost generalization. We examine how inductive bias affects attention-based models and discover that SegFormer performs better on bigger datasets. To further demonstrate the function of data augmentation, Low-Rank Adaptation (LoRA) is used to lower processing complexity while preserving accuracy. We show that the suggested Habaek model outperforms current models in segmentation, with an Intersection over Union (IoU) ranging from 0.91986 to 0.94397. In terms of F1-score, recall, accuracy, and precision, Habaek performs better than rival models, indicating its potential for real-world applications. This study highlights the need to enhance structures and include datasets for effective water segmentation.\end{abstract}


\section{Introduction}
Water segmentation and monitoring are vital for effective disaster response and water resource management \citep{loucks2017water}. Using high-resolution imagery, authorities can monitor rivers, lakes, and reservoirs to track changes over time \citep{marce2016automatic}. This enables proactive management of water supplies for agriculture, industry, and drinking purposes, as well as monitoring pollution sources and describing buildup to ensure water conservation.

Deep learning has significantly improved the efficiency of flood monitoring and prediction, making water segmentation faster and more accurate. Models like CNNs, U-Nets, and even transformers proved that they can handle large volumes of satellite and aerial imagery with precision, even in complex environments \citep{zhao2024spt}\citep{zhang2023mu}\citep{zhang2023actnet}. These advancements help automatically identify flood extents, providing valuable information to authorities during flooding events.

The integration of remote sensing technologies further enhances these deep learning models. Remote sensing ensures that accurate data is always available for flood monitoring, even in adverse weather conditions \citep{kussul2011flood}. Deep learning and remote sensing technologies work collaboratively to significantly improve flood monitoring and prediction. Several studies were also conducted, focusing on how models like U-Net, DeepLabv3+, and their variants perform segmentation tasks efficiently using satellite imagery \citep{wu2023near}\citep{yadav2022deep}. By using temporal data from remote sensors, these models can also predict how a flood will progress over time, enabling proactive measures to be taken \citep{schumann2009progress}. Together, these tools contribute to more reliable flood predictions and are essential for developing effective disaster response strategies, ultimately helping to minimize the impact of flooding on communities.

While deep learning models offer promising water segmentation solutions, they still have limitations. Despite being more accurate than traditional methods, these models often require significant computational resources, especially when handling high-resolution imagery and complex algorithms \citep{cheon2024combining}. This computational burden limits their predictability for real-time flood monitoring, posing challenges in balancing accuracy with efficiency \citep{jiang2022survey}. Therefore, the primary objective of this research is to develop a water segmentation model that achieves an optimal balance between accuracy and computational efficiency, ensuring its practical applicability for real-time use. The specific objectives of this study are as follows:

\begin{enumerate}
    \item Enhancing segmentation accuracy: Enhancing the Segformer to outperform existing methods while maintaining high IoU in water body segmentation tasks.
    
    \item Applying data augmentation: Incorporate diverse and relevant datasets, such as ADE20k, RIWA and Lufi-Riversnap dataset, to improve the model's generalization capability and validate on the Lufi-Riversnap dataset.
    
    \item Addressing the impact of inductive bias: Investigate the role of inductive bias in attention-based models like SegFormer, and demonstrate how larger datasets can promote its effects, leading to better segmentation outcomes.
\end{enumerate}

\section{Materials and Methods}
\subsection{Dataset Description and Preprocessing}

In this study, we utilized a total of three datasets for training and validation. The first dataset is available at: 
\href{https://www.kaggle.com/datasets/gvclsu/water-segmentation-dataset/data}{Water Segmentation Dataset}. 
From this dataset, we utilized the river-related train and validation images from the ADE20K dataset. 
The train set comprises 1,727 images, while the validation set contains 161 images. The second dataset, named \textit{River Water Segmentation (RIWA)}, focuses on the binary segmentation classification of river water and is available at 
\href{https://www.kaggle.com/datasets/franzwagner/river-water-segmentation-dataset}{RIWA Dataset}. It includes river photos captured by smartphones, drones, and DSLR cameras, along with water segmentation data and a subset of river-related images from ADE20K . 
We utilized RIWA version 2, which contains 1,142 training images and 167 validation images. The third dataset, \textit{Lufi-Riversnap}, is available at: 
\href{https://www.kaggle.com/datasets/arminmoghimi/lufi-riversnap}{Lufi-Riversnap Dataset}. It consists of river images taken by UAVs, surveillance cameras, smartphones, and regular cameras. To enhance the dataset’s diversity and accuracy, it also incorporates some images from both the first and second datasets. 
This dataset includes 657 training images, 202 validation images, and 233 test images. Since some datasets referenced the same sources, we conducted a cross-check for duplicate images between the train/validation and test sets. Any overlapping images were removed from the train/validation sets to ensure the integrity of the experiments.

\subsection{Segformer}
SegFormer, created by Xie et al., has two primary components: a hierarchical Transformer encoder and a lightweight all-MLP decoder. The hierarchical Transformer encoder pulls multi-scale features from the input picture without using positional encoding, which helps to prevent performance loss when test and training image resolutions change. The encoder produces multi-level features at various resolutions of the source picture, which are then utilized to increase segmentation accuracy. Unlike ViT, which employs 16x16 patches, SegFormer breaks the input picture into smaller 4x4 patches that are more suited to segmentation tasks. Furthermore, the encoder applies a sequence reduction method to reduce the complexity of self-attention, hence enhancing computational efficiency. SegFormer also uses overlapping patch merging to provide local continuity between patches. In contrast to non-overlapping techniques, the model provides overlapping sections that retain local context by specifying patch size, stride, and padding. The lightweight all-MLP decoder collects information from the encoder's multi-level features. It uses the hierarchical encoder's huge effective receptive field to combine information without the need for complicated handmade components. This design makes the decoder efficient while still providing powerful representations for segmentation \citep{xie2021segformer}.

\begin{figure}[h!]
	\centering
	\includegraphics[width=0.8\textwidth]{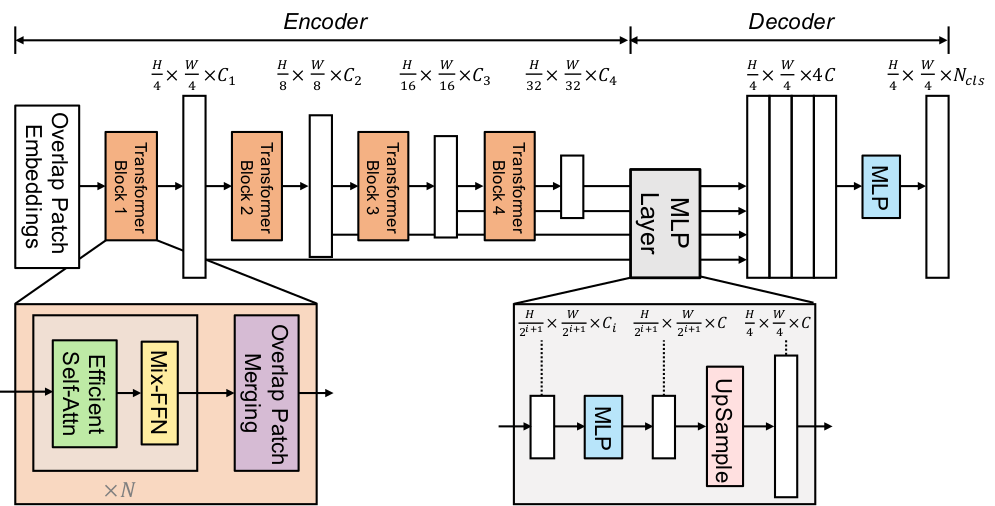}
	\caption{Architecture of SegFormer: Hierarchical transformer design for semantic segmentation}
	\label{fig:fig1}
\end{figure}

\subsection{Inductive Bias}
Indutive bias is the assumption that machine learning or deep learning models make about provided datasets. This inductive bias allows the models to effectively optimize from training data to unseen data. For example, CNN employs an inductive bias that local spatial patterns, like as edges and textures, are relevant in pictures. However, vision transformers (VITs) that use attention rather than convolution have less inductive bias. It suggests that ViTs use a global self-attention mechanism to record links between visual patches without making significant assumptions about the spatial localization of features. While this makes ViTs more flexible, they also need to learn these relationships from scratch, which requires a large amount of data to prevent overfitting and achieve satisfactory generalization \citep{zheng2024convolutional}. Dosovitskiy et al., compared ResNet to VIT on linear few-shot evaluation on ImageNet versus pre-training size. According to their experiments, CNNs (like R152x2 and R50x1) performed well with smaller datasets, while ViTs (like ViT-L-32 and ViT-B-32) require more data to shine, but they outperformed CNNs when enough data was available \citep{dosovitskiy2020image}. Since we aim to improve the performance of the pretrained SegFormer model, we opted to use a data augmentation technique. Given that SegFormer is built on the ViT technique, we believe that with a bigger, water-related dataset, SegFormer might achieve better performance without requiring extensive hyperparameter adjustment. This is consistent with the idea that ViT-based models gain considerably from bigger datasets since their lower inductive bias allows them to catch more complex patterns when given enough data.

\subsection{Low-Rank Adaptation: LoRA}
LoRA (Low-Rank Adaptation) advances neural networks by lowering dimensional complexity while preserving performance. Traditional neural networks utilize full-rank weight matrices, however, pretrained models frequently have a low intrinsic dimension, which means they may still learn effectively when transferred to smaller subspaces via random projections. The intrinsic dimension refers to the minimum number of dimensions required to express essential information, while random projection reduces dimensionality without losing significant data integrity.

LoRA applies these concepts by decomposing the weight matrix \(W_0\) (with dimensions \(d \times k\)) into two smaller matrices: \(B\) (dimensions \(d \times r\)) and \(A\) (dimensions \(r \times k\)), where \(r\) is much smaller than \(d\) and \(k\). During training, the original weight \(W_0\) remains fixed, and only the low-rank matrices \(A\) and \(B\) are updated. The resulting decomposition is expressed as:
\[W_0 + \Delta W = W_0 + BA\] where \(\Delta W\) is initially set to zero. LoRA further scales gradient updates by \(\alpha / r\), with \(\alpha\) acting as a constant similar to a learning rate. LoRA's parameter-efficient approach enables models to adapt to individual tasks while using few resources, making it particularly useful for transfer learning and segmentation tasks.

\section{Result}
In Table 1, we compared the average training time per epoch for various models on the LuFI dataset, along with their respective parameter counts. The Habaek model, which we proposed, stood out with a significantly reduced training time of 394 seconds per epoch, despite having 82.0M parameters, larger than most of the compared models. Other models like U-Net, PSPNet, DeepLabV3+, PAN, and LinkNet had training times ranging from 289 to 313 seconds per epoch, with parameter counts between 24M and 32M. Notably, SAM, the model with the highest number of parameters (94M), had a considerably longer training time of 424 seconds per epoch. In contrast, Habaek achieved an impressive balance between parameter size and efficiency, showcasing its scalability and faster training on the LuFI dataset. This result demonstrated that Habaek could be an excellent choice for real-world applications requiring fast and accurate water body segmentation. 

\begin{table}[ht]
\centering
\caption{Comparison of Training Time and Parameters for Various Models on the LuFI Dataset}
\begin{tabular}{lcc}
\toprule
\textbf{Model} & \textbf{Training Time (seconds/epoch)} & \textbf{Parameter Count (M)} \\
\midrule
U-Net        & \textbf{289} & \textbf{24} \\
PSPNet       & 305 & 30 \\
DeepLabV3+   & 313 & 32 \\
PAN          & 295 & 28 \\
LinkNet      & 298 & 26 \\
SAM          & 424 & 94 \\
Habaek  & 394 & {82} \\
\bottomrule
\end{tabular}
\label{tab:lufi_training_comparison}
\end{table}

Table 2 presents the test Intersection over Union (IoU) results for models trained and validated on various dataset combinations, evaluated on the LuFI test set. In the first experiment, the model was trained and validated solely on the LuFI dataset, achieving a baseline IoU of 0.91986. In the second experiment, where both the LuFI and ADE20K datasets were used for training and validation, the IoU improved significantly to 0.93677, suggesting that incorporating additional, external data helped the model generalize better to the target test set. Finally, in the third setup, the training and validation datasets included LuFI, ADE20K, and RIWA, resulting in the highest IoU of 0.94397. This demonstrated that combining diverse water-related datasets provided complementary information, enhancing the model's ability to capture more diverse features and improve segmentation performance. Those performances were measured by:

\begin{align}
    OA &= 1 - \frac{TP + TN}{TP + FP + FN + TN} \times 100\% \tag{1} \\
    Precision &= \frac{TP}{TP + FP} \times 100\% \tag{2} \\
    Recall &= \frac{TP}{TP + FN} \times 100\% \tag{3} \\
    F_S &= \frac{2TP}{2TP + FP + FN} \times 100\% \tag{4} \\
    IoU &= \frac{TP}{TP + FP + FN} \times 100\% \tag{5}
\end{align}

\begin{table}[ht]
\centering
\caption{Test IoU Results for Models Trained on Various Dataset Combinations Evaluated on the LuFI Test Set}
\begin{tabular}{lcc}
\toprule
\textbf{experiment} & \textbf{Training and Validation Datasets} & \textbf{IoU} \\
\midrule
Experiment 1 & LuFI only & 0.91986 \\
Experiment 2 & LuFI + ADE20K & 0.93677 \\
Experiment 3 & LuFI + ADE20K + RIWA & \textbf{0.94397} \\
\bottomrule
\end{tabular}
\label{tab:iou_results}
\end{table}

Table 3 also compared various segmentation models based on their performance metrics: Overall Accuracy (OA), IoU, Precision, Recall, and the F1-score. The Habaek model consistently outperformed the other models across all metrics. It achieved the highest OA of 0.9754, significantly surpassing the second-best model, PAN, which achieved an OA of 0.966. In terms of IoU, a crucial metric for segmentation tasks, Habaek again took the lead with a score of 0.94397, compared to 0.925 from SAM, the second-best model in this category. When examining Precision, SAM achieved the highest value at 0.985, while Habaek followed closely at 0.9748. However, in terms of Recall, Habaek dominated with a score of 0.9671, compared to 0.942 from U-Net and DeepLabV3+. Similarly, the F1-score, which balances Precision and Recall, was highest for Habaek at 0.9709, confirming its performance across multiple dimensions. These results highlighted that Habaek excelled in segmenting target features, proving its effectiveness in a variety of segmentation tasks. The model’s consistent top performance across OA, IoU, Recall, and F1-score underscored its potential as a strong alternative to traditional models for segmentation applications.

\begin{table}[ht]
\centering
\caption{Performance comparison of different models}
\label{tab:my-table}
\begin{tabular}{|l|c|c|c|c|c|}
\hline
\multirow{2}{*}{\textbf{Model}} & \multicolumn{5}{c|}{\textbf{Metrics}} \\ \cline{2-6} 
                                & \textbf{OA} & \textbf{IoU} & \textbf{Precision} & \textbf{Recall} & \textbf{$F_s$} \\ \hline
U-Net(\textit{ResNet50})        & 0.961       & 0.890       & 0.939             & 0.942             & 0.934         \\ \hline
PSPNet(\textit{ResNet50})       & 0.941       & 0.833       & 0.951             & 0.865             & 0.899         \\ \hline
DeeplabV3+(\textit{ResNet50})   & 0.961       & 0.898       & 0.951             & 0.938             & 0.941         \\ \hline
PAN(\textit{ResNet50})          & 0.966       & 0.892       & 0.952             & 0.938             & 0.938         \\ \hline
LinkNet(\textit{ResNet50})      & 0.958       & 0.884       & 0.939             & 0.932             & 0.930         \\ \hline
SAM                             & 0.963       & 0.925       & \textbf{0.985}    & 0.938             & 0.957         \\ \hline
Habaek                     & \textbf{0.9754} & \textbf{0.9439} & 0.9748             & \textbf{0.9671}    & \textbf{0.9709} \\ \hline
\end{tabular}
\end{table}

The evaluation results further demonstrated that using LoRA (Low-Rank Adaptation) with a larger dataset, even without full fine-tuning, yielded superior segmentation performance. During LoRA training, the model achieved a notable IoU of 0.9387, alongside high accuracy (0.9730), Precision (0.9711), Recall (0.9653), and F1-score (0.9681). These results were significantly better than the baseline IoU of 0.91986, achieved when the model was trained solely on the LuFI dataset. This indicated that expanding the dataset, even without full fine-tuning, markedly improved model performance. Except for our proposed model, the other results were obtained from the research of Moghimi et al. \citep{moghimi2024comparative}.

Figures 2 and 3 display the segmentation performance on the LuFi dataset, highlighting the model’s ability to detect diverse water features. These examples illustrate the model’s versatility in segmenting both large and small water features. Figure 4 demonstrates an overlay of predicted water regions (in blue) on the original scene, including a river beneath a bridge. Areas highlighted in green represent over-predictions, where the model incorrectly identifies non-water areas as water. In contrast, areas marked in red indicate missed predictions, where the model failed to detect water. 
\begin{table}[ht]
\centering
\caption{Test IoU Results for Different Dataset Configurations and LoRA Training on the LuFI Test Set}
\begin{tabular}{lcc}
\toprule
\textbf{Configuration} & \textbf{Datasets} & \textbf{IoU} \\
\midrule
Baseline & LuFI only & 0.91986 \\
Dataset Expansion & LuFI + ADE20K & 0.93677 \\
Full Dataset & LuFI + ADE20K + RIWA & \textbf{0.94397} \\
LoRA Training & LuFI + ADE20K + RIWA & 0.9387 \\
\bottomrule
\end{tabular}
\label{tab:iou_results}
\end{table}

\begin{figure}[h!]
    \centering
    \includegraphics[width=0.8\textwidth]{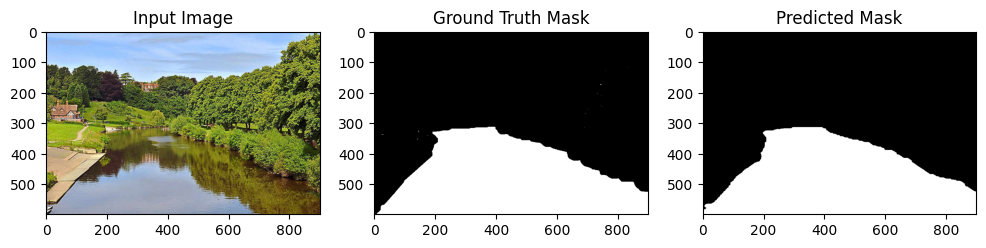}
    \caption{Segmentation results showing river detection on the LuFi dataset.}
    \label{fig:fig1}
\end{figure}

\begin{figure}[h!]
    \centering
    \includegraphics[width=0.8\textwidth]{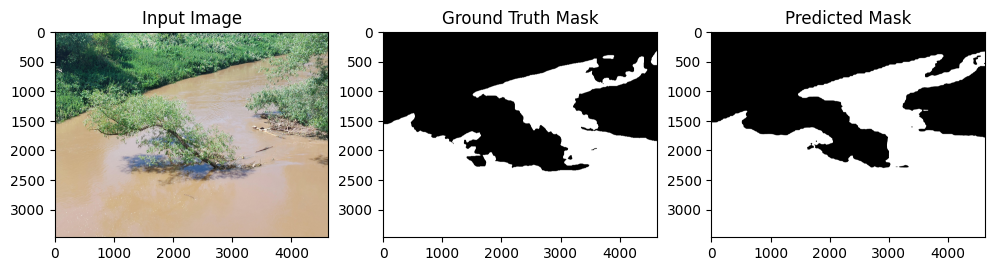}
    \caption{Segmentation results highlighting small water bodies and streams on the LuFi dataset.}
    \label{fig:fig2}
\end{figure}

\begin{figure}[h!]
    \centering
    \includegraphics[width=0.8\textwidth]{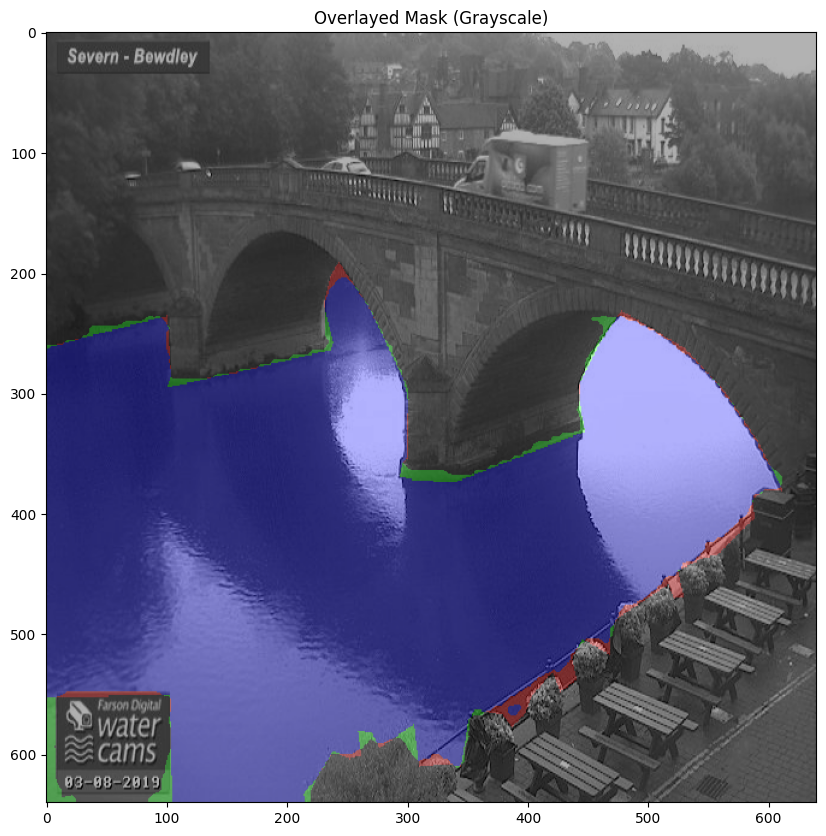}
    \caption{Segmentation performance comparison on a complex terrain with water bodies in the LuFi dataset.}
    \label{fig:fig3}
\end{figure}

\section{Conclusion}
This work conducted several tests to improve the performance of deep learning-based water segmentation. For the experiment, we chose the SegFormer model, which has a lightweight design in comparison to many other models. The essence of our suggested strategy is based on utilizing the inductive bias of the given model. We set a hypothesis that since the SegFormer uses an attention mechanism, which implies lower inductive bias could be enhanced with larger datasets.  We used data augmentation, including ADE20k and RIWA in the training process to prove our hypothesis. Habaek, our proposed model, demonstrates exceptional performance and efficiency, establishing itself as a dependable choice for water body segmentation tasks. Habaek requires less training time per epoch than other models, even though it consists of more parameters. With the Intersection over Union (IoU) rising from 0.91986 to 0.94278, the segmentation accuracy was significantly improved by including ADE20k and RIWA datasets. Furthermore, using LoRA adaptation on bigger datasets improved model performance efficiently. Habaek outperformed other models on important measures like as accuracy, precision, recall, and F1-score, demonstrating its efficacy for real-world applications. To sum up, this paper proves that Habaek performs at state-of-the-art (SOTA) levels in water segmentation tasks. This result also underscores the necessity of data augmentation, particularly for models with low inductive bias.

\bibliographystyle{unsrtnat}
\bibliography{template}

\begin{thebibliography}{15}
\providecommand{\natexlab}[1]{#1}
\providecommand{\url}[1]{\texttt{#1}}
\expandafter\ifx\csname urlstyle\endcsname\relax
  \providecommand{\doi}[1]{doi: #1}\else
  \providecommand{\doi}{doi: \begingroup \urlstyle{rm}\Url}\fi

\bibitem[Loucks and Van~Beek(2017)]{loucks2017water}
Daniel~P Loucks and Eelco Van~Beek.
\newblock \emph{Water resource systems planning and management: An introduction to methods, models, and applications}.
\newblock Springer, 2017.

\bibitem[Marc{\'e} et~al.(2016)Marc{\'e}, George, Buscarinu, Deidda, Dunalska, de~Eyto, Flaim, Grossart, Istvanovics, Lenhardt, et~al.]{marce2016automatic}
Rafael Marc{\'e}, Glen George, Paola Buscarinu, Melania Deidda, Julita Dunalska, Elvira de~Eyto, Giovanna Flaim, Hans-Peter Grossart, Vera Istvanovics, Mirjana Lenhardt, et~al.
\newblock Automatic high frequency monitoring for improved lake and reservoir management.
\newblock \emph{Environmental Science \& Technology}, 50\penalty0 (20):\penalty0 10780--10794, 2016.

\bibitem[Zhao et~al.(2024)Zhao, Du, Xu, Jian, Pei, Zhu, Yan, and Fan]{zhao2024spt}
Teng Zhao, Xiaoping Du, Chen Xu, Hongdeng Jian, Zhipeng Pei, Junjie Zhu, Zhenzhen Yan, and Xiangtao Fan.
\newblock Spt-unet: A superpixel-level feature fusion network for water extraction from sar imagery.
\newblock \emph{Remote Sensing}, 16\penalty0 (14):\penalty0 2636, 2024.

\bibitem[Zhang et~al.(2023{\natexlab{a}})Zhang, Lu, Ma, Zhao, Xie, Geng, Tian, and Sian]{zhang2023mu}
Yonghong Zhang, Huanyu Lu, Guangyi Ma, Huajun Zhao, Donglin Xie, Sutong Geng, Wei Tian, and Kenny Thiam Choy Lim~Kam Sian.
\newblock Mu-net: Embedding mixformer into unet to extract water bodies from remote sensing images.
\newblock \emph{Remote Sensing}, 15\penalty0 (14):\penalty0 3559, 2023{\natexlab{a}}.

\bibitem[Zhang et~al.(2023{\natexlab{b}})Zhang, Liu, Liu, Tian, and Qu]{zhang2023actnet}
Zheng Zhang, Fanchen Liu, Changan Liu, Qing Tian, and Hongquan Qu.
\newblock Actnet: A dual-attention adapter with a cnn-transformer network for the semantic segmentation of remote sensing imagery.
\newblock \emph{Remote Sensing}, 15\penalty0 (9):\penalty0 2363, 2023{\natexlab{b}}.

\bibitem[Kussul et~al.(2011)Kussul, Shelestov, and Skakun]{kussul2011flood}
Nataliia Kussul, Andrii Shelestov, and Sergii Skakun.
\newblock Flood monitoring from sar data.
\newblock In \emph{Use of satellite and in-situ data to improve sustainability}, pages 19--29. Springer, 2011.

\bibitem[Wu et~al.(2023)Wu, Zhang, Xiong, Zhang, Tang, Li, An, and Li]{wu2023near}
Xuan Wu, Zhijie Zhang, Shengqing Xiong, Wanchang Zhang, Jiakui Tang, Zhenghao Li, Bangsheng An, and Rui Li.
\newblock A near-real-time flood detection method based on deep learning and sar images.
\newblock \emph{Remote Sensing}, 15\penalty0 (8):\penalty0 2046, 2023.

\bibitem[Yadav et~al.(2022)Yadav, Nascetti, and Ban]{yadav2022deep}
Ritu Yadav, Andrea Nascetti, and Yifang Ban.
\newblock Deep attentive fusion network for flood detection on uni-temporal sentinel-1 data.
\newblock \emph{Frontiers in Remote Sensing}, 3:\penalty0 1060144, 2022.

\bibitem[Schumann et~al.(2009)Schumann, Bates, Horritt, Matgen, and Pappenberger]{schumann2009progress}
Guy Schumann, Paul~D Bates, Matthew~S Horritt, Patrick Matgen, and Florian Pappenberger.
\newblock Progress in integration of remote sensing--derived flood extent and stage data and hydraulic models.
\newblock \emph{Reviews of Geophysics}, 47\penalty0 (4), 2009.

\bibitem[Cheon and Mun(2024)]{cheon2024combining}
Minjong Cheon and Changbae Mun.
\newblock Combining kan with cnn: Konvnext’s performance in remote sensing and patent insights.
\newblock \emph{Remote Sensing}, 16\penalty0 (18):\penalty0 3417, 2024.

\bibitem[Jiang et~al.(2022)Jiang, Peng, Zhong, Xie, Hao, Lin, Ma, and Hu]{jiang2022survey}
Huiwei Jiang, Min Peng, Yuanjun Zhong, Haofeng Xie, Zemin Hao, Jingming Lin, Xiaoli Ma, and Xiangyun Hu.
\newblock A survey on deep learning-based change detection from high-resolution remote sensing images.
\newblock \emph{Remote Sensing}, 14\penalty0 (7):\penalty0 1552, 2022.

\bibitem[Xie et~al.(2021)Xie, Wang, Yu, Anandkumar, Alvarez, and Luo]{xie2021segformer}
Enze Xie, Wenhai Wang, Zhiding Yu, Anima Anandkumar, Jose~M Alvarez, and Ping Luo.
\newblock Segformer: Simple and efficient design for semantic segmentation with transformers.
\newblock \emph{Advances in neural information processing systems}, 34:\penalty0 12077--12090, 2021.

\bibitem[Zheng et~al.(2024)Zheng, Li, and Lucey]{zheng2024convolutional}
Jianqiao Zheng, Xueqian Li, and Simon Lucey.
\newblock Convolutional initialization for data-efficient vision transformers.
\newblock \emph{arXiv preprint arXiv:2401.12511}, 2024.

\bibitem[Dosovitskiy(2020)]{dosovitskiy2020image}
Alexey Dosovitskiy.
\newblock An image is worth 16x16 words: Transformers for image recognition at scale.
\newblock \emph{arXiv preprint arXiv:2010.11929}, 2020.

\bibitem[Moghimi et~al.(2024)Moghimi, Welzel, Celik, and Schlurmann]{moghimi2024comparative}
Armin Moghimi, Mario Welzel, Turgay Celik, and Torsten Schlurmann.
\newblock A comparative performance analysis of popular deep learning models and segment anything model (sam) for river water segmentation in close-range remote sensing imagery.
\newblock \emph{IEEE Access}, 2024.

\end{thebibliography}
\end{document}